\documentclass[10pt,twocolumn,letterpaper]{article}

\usepackage{iccv}
\usepackage{times}
\usepackage{epsfig}
\usepackage{graphicx}
\usepackage{amsmath}
\usepackage{amssymb}
\usepackage{tabularray}
\usepackage{verbatim}
\usepackage[normalem]{ulem}
\useunder{\uline}{\ul}{}
\usepackage{pifont}
\usepackage{booktabs}

\usepackage{enumitem}
\usepackage{placeins}
\usepackage{xcolor}
\usepackage{flushend}

\usepackage[pagebackref=true,breaklinks=true,letterpaper=true,colorlinks,bookmarks=false,citecolor=blue]{hyperref}

\newcommand{\myparagraph}[1]{\smallskip\noindent\textbf{#1}}

\newcommand{\cmark}{\ding{51}}%
\newcommand{\xmark}{\ding{55}}%
\newcommand{\x}{\mathbf{x}}

\newcommand{\xdd}{\x^{\text{2D}}}
\newcommand{\xtd}{\x^{\text{3D}}}
\newcommand{\edd}{\mathbf{E}^{\text{2D}}}
\newcommand{\etd}{\mathbf{E}^{\text{3D}}}

\newcommand{\I}{\mathbf{I}}
\newcommand{\E}[1]{\mathbb{E} [ #1 ]}

\newcommand{\f}[1]{\mathrm{#1}}
\newcommand{\R}{\mathbb{R}}
\newcommand{\N}{\mathcal{N}}
\newcommand{\denoiser}{\epsilon_\theta}
\newcommand{\enc}{\f{w}_\theta}

\newcommand{\CR}[1]{\textcolor{olive}{#1}}

\iccvfinalcopy 

\def\iccvPaperID{****} 

\ificcvfinal\fi

\begin{document}

\title{DiffHPE: Robust, Coherent 3D Human Pose Lifting with Diffusion}

\author{
Cédric Rommel$^1$\\
\and
Eduardo Valle$^{1,4}$\\  
\and
Mickaël Chen$^1$\\
\and
Souhaiel Khalfaoui$^2$\\
\and
Renaud Marlet$^{1,5}$\\
\and
Matthieu Cord$^{1,3}$\\
\and
Patrick Pérez$^1$ \\
\and
$^1$Valeo.ai, Paris, France \;
$^2$Valeo - Applied ML, Paris, France \;
$^3$Sorbonne Université, Paris, France \\
$^4$Recod.ai Lab, School of Electrical and Computing Engineering, University of Campinas, Brazil \\
$^5$LIGM, Ecole des Ponts, Univ Gustave Eiffel, CNRS, Marne-la-Vallee, France
}

\maketitle
\ificcvfinal\thispagestyle{empty}\fi

\begin{abstract}
    We present an innovative approach to 3D Human Pose Estimation (3D-HPE) by integrating cutting-edge diffusion models, which have revolutionized diverse fields, but are relatively unexplored in 3D-HPE.  We show that diffusion models enhance the accuracy, robustness, and coherence of human pose estimations.
    We introduce DiffHPE, a novel strategy for harnessing diffusion models in 3D-HPE, and demonstrate its ability to refine standard supervised 3D-HPE. We also show how diffusion models lead to more robust estimations in the face of occlusions, and improve the time-coherence and the sagittal symmetry of predictions. Using the Human\,3.6M dataset, we illustrate the effectiveness of our approach and its superiority over existing models, even under adverse situations where the occlusion patterns in training do not match those in inference. Our findings indicate that while standalone diffusion models provide commendable performance, their accuracy is even better in combination with supervised models, opening exciting new avenues for 3D-HPE research.
\end{abstract}

\section{Introduction}

3D Human Pose Estimation (3D-HPE) is rapidly evolving, with recent methods fast advancing in accuracy~\cite{gong_diffpose_2022, shan2023diffusion}. This work joins those efforts, showcasing how integrating diffusion models into state-of-the-art models enhances
not only their accuracy as previously understood, but also their robustness and coherence, as we demonstrate in our experiments.

Diffusion models, a cutting-edge generative technique, are making waves across various domains, including computer vision~\cite{nichol2021glide, rombach2022high, saharia2022photorealistic, ramesh2022hierarchical,lugmayr2022repaint,saharia2022palette}, natural language processing~\cite{austin2021structured,li2022diffusion,campbell2022continuous} and time-series analysis~\cite{kong2020diffwave, tashiro_csdi_2021,saadatnejad_generic_2022}.

The application of diffusion models to 3D-HPE (and other purely predictive tasks) remains largely unexplored, despite having shown remarkable performance in human pose forecasting, including strong robustness to occlusions~\cite{saadatnejad_generic_2022}. In this work, we show how their ability to overcome ambiguities transposes from the latter to the former.

While a few pioneering works have shown promising performance metrics~\cite{gong_diffpose_2022, shan2023diffusion}, the understanding of the benefits of diffusion models over classical supervision --- as well as key design choices --- is still in its infancy. In this work, we address those concerns, providing an in-depth analysis of the effects of diffusion models on 3D-HPE. 

Our contributions are threefold:

\begin{enumerate}[itemsep=1mm]
\item We propose \mbox{\emph{DiffHPE}}, a novel strategy to use diffusion models in 3D-HPE;
\item We show that combining diffusion with supervised 3D-HPE (\mbox{\emph{DiffHPE-Wrapper}}) outperforms each model trained separately;
\item Our extensive analyses showcase how diffusion models' estimations display better bilateral and temporal coherence, and are more robust to occlusions, even when not perfectly trained for the latter.
\end{enumerate}

\section{Related work}


\myparagraph{3D Human pose estimation.} 
Compared to 2D, 3D-HPE is much less mature, with more incipient results, especially in \emph{monocular} 3D-HPE, which will be our scope. Early works in monocular 3D-HPE tackled the problem end-to-end, using deep neural networks to predict 3D keypoints directly from images \cite{pavlakos_coarse--fine_2017, moreno20173d, mehta2017vnect, sun_compositional_2017}. Due to persistent difficulties faced by that approach and much faster advances in 2D-HPE, the current state of the art employs a 2-step pipeline, first applying 2D-HPE, and then \textit{lifting} the 2D results into 3D space.

The first methods for pose lifting used small multi-layer perceptrons~\cite{martinez_simple_2017} and nearest-neighbors matching~\cite{chen_3d_2017}.

While human pose estimation was initially tackled at the frame level, the field quickly adopted recurrent~\cite{hossain_exploiting_2018} and convolutional neural networks~\cite{pavllo_3d_2019} to move towards video-level predictions. That allowed leveraging temporal correlations to improve accuracies. Graph convolutional networks were then proposed to exploit the keypoints' connectivity, drastically reducing the computational complexity while improving results~\cite{zhao_semantic_2019, liu_comprehensive_2020, cai_exploiting_2019, zou_modulated_2021, hu_conditional_2021, xu_graph_2021}.

More recently, spatial-temporal transformer architectures were proposed~\cite{shan_p-stmo_2022, zheng_3d_2021}, including MixSTE~\cite{zhang_mixste_2022}, which, arguably, is the state of the art for 3D human pose lifting among deterministic methods.

\myparagraph{Generative human pose estimation.}
Lifting human pose to 3D is an inherently ambiguous task since many 3D poses may project onto the same 2D input.
That led the community to investigate multi-hypothesis approaches based on generative models, such as variational autoencoders~\cite{sharma2019monocular}, normalizing flows~\cite{kolotouros2021probabilistic,wehrbein2021probabilistic} and, more recently, diffusion models~\cite{gong_diffpose_2022, shan2023diffusion}.

DiffPose~\cite{gong_diffpose_2022} and D3DP~\cite{shan2023diffusion} employ a denoiser based on MixSTE, trained from scratch. D3DP~\cite{shan2023diffusion} conditions the diffusion on the raw 2D keypoints, in a scheme that has parallels to our DiffHPE-2D (\autoref{sec:diffwrap-method}).
It introduces a novel hypotheses-aggregation scheme, more sophisticated than simple averaging, based on 2D reprojections, but which depends on the availability of the camera parameters.

DiffPose~\cite{gong_diffpose_2022} recently set a new state of the art in 3D human pose estimation. 
It employs an unusual diffusion procedure based on Gaussian mixture models learned on 2D heatmaps created from the predictions of the upstream models. Besides having MixSTE as a denoiser, it employs a pre-trained MixSTE to initialize the reverse diffusion during inference --- at least in the frame-level model, which is the only one released at this time.\footnote{\url{https://github.com/GONGJIA0208/Diffpose/blob/af2954513f6f5df274466bf4a45fb84c588b48c6/runners/diffpose_frame.py}}

Our proposal diverges considerably from those works. Our denoiser architecture is based on CSDI~\cite{tashiro_csdi_2021} and TCD~\cite{saadatnejad_generic_2022} but introduces graph-convolutional layers that allow for good accuracy, with less computational burden than CSDI transformers. We employ a streamlined standard diffusion that forgoes the complexities of DiffPose. We use a frozen, pre-trained MixSTE as conditioning of the diffusion process, during both training and inference.


\myparagraph{Denoising Diffusion Probabilistic Models.}
Denoising diffusion probabilistic models (DDPM)~\cite{ho_denoising_2020} emerged as the new state-of-the-art generative models~\cite{yang2022diffusion}, leading to impressive results in a broad range of applications such as text-to-image generation~\cite{nichol2021glide, rombach2022high, saharia2022photorealistic, ramesh2022hierarchical}, inpainting~\cite{lugmayr2022repaint, saharia2022palette}, audio synthesis~\cite{kong2020diffwave}, time-series imputation~\cite{tashiro_csdi_2021} and computational chemistry~\cite{hoogeboom_equivariant_2022}.

They were recently applied to 3D human pose forecasting~\cite{saadatnejad_generic_2022},
showing state-of-the-art performance, including in scenarios with strong occlusions.
We took great inspiration from this work, pursuing similar occlusion robustness for 3D human pose estimation.

Both our architecture and TCD~\cite{saadatnejad_generic_2022} are based on CSDI~\cite{tashiro_csdi_2021}, but we exchange the computation-intensive transformers of CSDI for graph-convolutional layers. We considerably extend the occlusion analysis of~\cite{saadatnejad_generic_2022} to contemplate distribution shifts between the occlusions observed in training and those found during inference. We also include a novel analysis of the effects of diffusion on the coherence of poses.

To our knowledge, this is the first study to observe the positive impacts of diffusion --- robustness to distribution shifts and improvement of symmetry and time-coherence --- on 3D human poses.

\section{Method}

\begin{figure*}[htp]
    \centering
    \includegraphics[width=\textwidth]{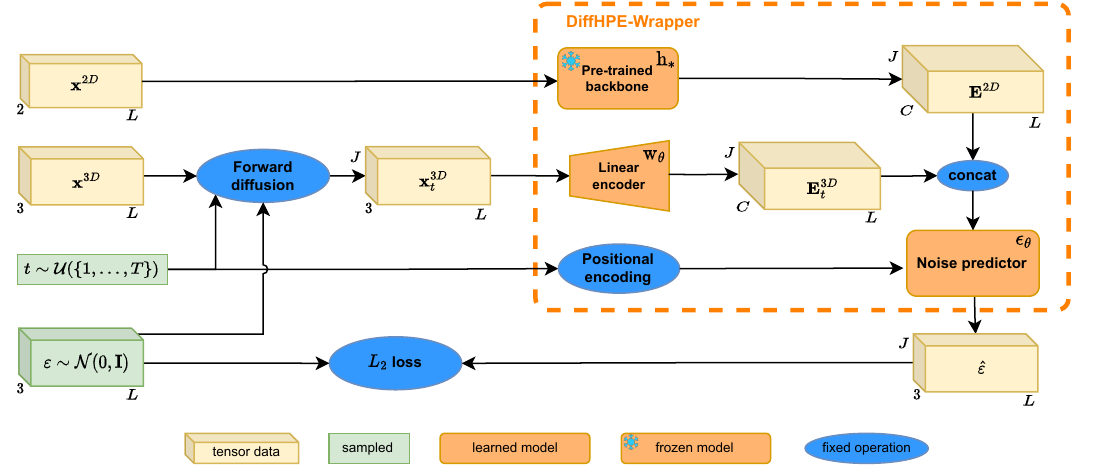}
    \caption{\textbf{Training procedure of the proposed diffusion-based 2D-to-3D human pose estimation lifting}. DiffHPE-Wrapper is composed of a frozen pre-trained lifting backbone $\f{h}_*$, a linear encoding layer $\enc$ for the 3D data, and noise-predicting deep neural network $\denoiser$. More details in subsections \ref{sec:diffwrap-method} and \ref{sec:training}.
    `Forward diffusion' corresponds to equation \eqref{eq:fwd-proc2} and `concat' to channel-wise concatenation. }
    \label{fig:training}
\end{figure*}

\subsection{Background on diffusion models} 

Our work builds upon DDPMs~\cite{ho_denoising_2020}, a unique approach to generative modeling that has gained enormous popularity due to its extraordinary ability to represent intricate distributions, while being stable to train (when compared to GANs) and allowing for flexible architectures (when compared to normalizing flows).

DDPMs are trained to reverse a diffusion process that gradually adds Gaussian noise to the training data $\x_0$ over $T$ steps until it becomes pure noise at $\x_T$. This \emph{forward diffusion process} may be formalized as sampling from a conditional distribution $\f{q}$
\begin{gather}
    \f{q}(\x_t| \x_{t-1}) = \N(\x_t; \sqrt{1 - \beta_t} \x_{t-1}, \beta_t \I) \,, \label{eq:fwd-proc}
\end{gather}
where $\I$ is the identity matrix and the rate at which the original data is diffused into noise is controlled by a \emph{variance scheduling} given by $\beta_1, \dots, \beta_T$.

Importantly, the forward process and its schedule may be reparameterized as
\begin{gather}
    \alpha_t = \prod_{t=1}^T (1 - \beta_t) ,\, \label{eq:fwd-alpha} \\
    \f{q}(\x_t| \x_0) = \N(\x_t; \sqrt{\alpha_t} \x_{t-1}, (1 -\alpha_t) \I), \label{eq:fwd-proc2}
\end{gather}
\noindent allowing to sample from each step $t$ directly. This ability is crucial for efficiently training the models.

If we knew $\f{p}(\x_{t-1}| \x_t)$, we could reverse the noise process, turning a Gaussian sample back into a sample from the data distribution. This reverse conditional distribution is arbitrarily complex, but
we might approximate it
via a denoising deep neural network. In particular, for small enough denoising steps, we may set
\begin{gather}
    \f{p}(\x_{t-1} | \x_t) \approx \N(\x_{t-1} ; \mu_\theta(\x_t, t), \Sigma_t) \,, \label{eq:denoising-posterior}
\end{gather}
where $\mu_\theta(\x_t, t)$ is a learned deep neural network that has as input both the noisy data $\x_t$ and its step $t$ (usually position-encoded); $\Sigma_t$ depends on the variance schedule but is not otherwise learned.

In practice, instead of modeling $\mu_\theta(\x_t, t)$ directly, one often prefers to, equivalently, infer the noise added to $\x_t$:
\begin{equation}
    \mu_\theta(\x_t, t) = k_{1,t}(\x_t - k_{2,t} \denoiser(\x_t, t)) \,,\label{eq:ddpm-mean}
\end{equation}
\noindent where $\denoiser$ is the noise-predicting neural network, and $k_{.,t}$ are constants that depend on the variance schedule terms.

Conditional diffusion models, which add an extra term in the denoising process, provide a window of opportunity for purely predictive tasks:
\begin{gather}
    \f{p}(\x_{t-1}|\x_t, c) \approx \N(\x_{t-1} ; \mu_\theta(\x_t, c, t), \Sigma_t) \,, \label{eq:ddpm-cond} \\
    \mu_\theta(\x_t, c, t) = k_{1,t}(\x_t - k_{2,t}\denoiser(\x_t, c, t)) \,,
\end{gather}
\noindent where the conditioning inputs $c$ added to $\denoiser$ may be derived from the predictive task inputs. Conditioning allows us to leverage the power of diffusion models for 3D-HPE lifting.

\subsection{Wrapping 2D-to-3D lifting with diffusion}
\label{sec:diffwrap-method}

As mentioned, the lifting approach to 3D-HPE consists in predicting the 3D positions $\xtd$ of human keypoints from corresponding 2D positions $\xdd$ pre-obtained by some upstream procedure. The $\xtd$ are in the reference frame of the camera, while the $\xdd$ lie in the pixel space of an image of height $h$ and width $w$.

We assume that 2D and 3D poses are available in a sequence of $L$ frames, each containing $J$ joints, such that $\xdd \in (\{1, \dots, h\} \times \{1, \dots, w\})^{L \times J} \subset \R^{L \times J \times 2}$ and $\xtd \in \R^{L \times J \times 3}$.

Traditional lifting amounts to training a deterministic, supervised regression model ${f_\theta: \R^{L \times J \times 2} \to \R^{L \times J \times 3}}$ on a dataset containing $N$ pairs ${\{(\xdd_i, \xtd_i): 1 \leq i \leq N\}}$, estimating the conditional expectation 
$\E{\xtd|\xdd}$,
typically using a standard mean-squared loss.

The generative flavor of lifting learns from the dataset not just the expectation, but the whole conditional distribution
$\f{p}(\xtd|\xdd)$, from which candidate 3D poses can be sampled. We propose using a diffusion model, with the 3D keypoints $\xtd$ as the target for the denoising process, and the 2D keypoints $\xdd$ as conditioners on the process, such that the conditional DDPM formalism leads to:
\begin{gather}
    \f{p}(\xtd_T|\xdd) = \N(\xtd_T; 0, \I) \,,\\
    \f{p}(\xtd_{t-1} | \xtd_t, \xdd) \approx \N(\xtd_{t-1} ; \mu_\theta(\xtd_t, \xdd, t), \Sigma_t) \,. \label{eq:ddpm-cond-2d}
\end{gather}

\myparagraph{Conditioning on the input 2D pixel coordinates.}
The straightforward diffusion model for lifting conditions the denoising of $\xtd$ directly on the raw 2D data $\xdd$, as shown in eq.~\eqref{eq:ddpm-cond-2d}. That is achieved in practice by feeding both $\xtd_t$ and $\xdd$ to the denoising network $\denoiser$ at every diffusion step $t$. We call this strategy \textit{DiffHPE-2D}.

\myparagraph{Conditioning on the features of a pre-trained model.}
An alternative to conditioning on the raw 2D inputs is leveraging the internal representations (feature vectors) of a pre-trained deterministic lifting model.

Suppose that ${\f{f}_* = \f{g}_* \circ \f{h}_*}$ is a frozen, pre-trained lifting neural network composed of a backbone ${\f{h}_*:\R^{L \times J \times 2} \to \R^{L \times J \times C}}$ and a linear regression head ${\f{g}_*: \R^{L \times J \times C} \to \R^{L \times J \times 3}}$.

\begin{figure}
    \centering
    \includegraphics[width=\columnwidth]{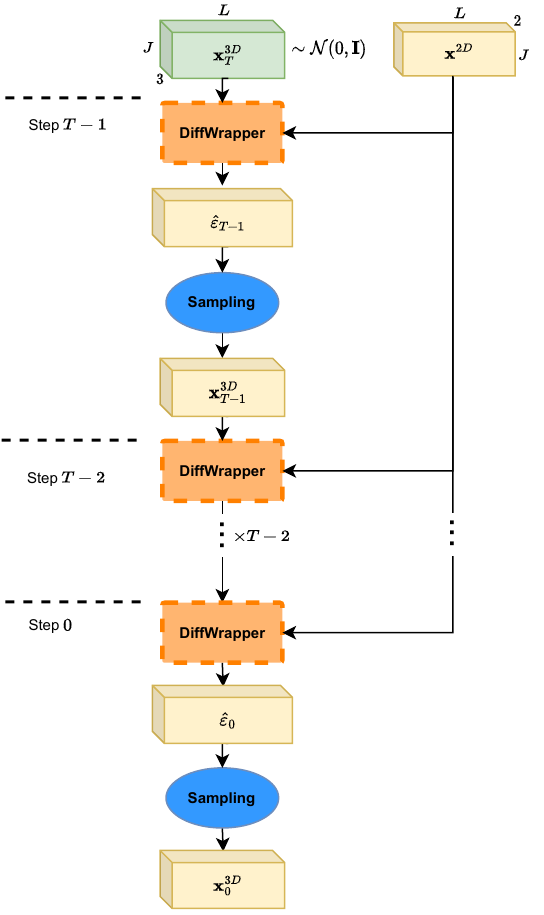}
    \caption{\textbf{Human pose lifting by diffusion on 3D pose, conditioned on 2D pose}. Sampling is done by following the reverse process where the noise is estimated by the trained DiffHPE model.
    }
    \label{fig:inference}
\end{figure}

We could use the output of $\f{h}_*$ instead of $\xdd$ in eq.~\eqref{eq:ddpm-cond-2d}. This rich feature vector has much more information than the 3D output of $\f{g}_*$. However, since generally $C \gg 3$, the conditioner overpowers $\xtd_t$ in $\mu_\theta$, leading to poor performance. We mitigate that issue by also embedding $\xtd_t$ into a higher-dimensional layer, such that the diffusion scheme becomes:
\begin{gather}
    \edd = \f{h}_*(\xdd) \,,\,\,
    \etd_t = \enc(\xtd_t) \,,\\
    \f{p}(\xtd_{t-1} | \xtd_t, \xdd) \approx \N(\xtd_{t-1} ; \mu_\theta(\etd_t, \edd, t), \Sigma_t) \,,\\
    \mu_\theta(\etd_t, \edd, t) = k_{1,t} (\xtd_t - k_{2,t} \denoiser(\etd_t, \edd, t)) \,, \label{eq:ddpm-cond-mixste}
\end{gather}
\noindent where $\enc(\xtd)$ is a single trainable linear layer, and $\denoiser$ is the trainable noise-predicting model for the DDPM. The whole scheme is illustrated in Figures \ref{fig:training}-\ref{fig:inference}.

Since this strategy can be seen as wrapping the original lifting model $\f{f}_*$ with the diffusion model, we
call it \mbox{\textit{DiffHPE-Wrapper}}.
This diffusion model focuses on refining the lifting without the burden of working directly on the raw $\xdd$ inputs.

\subsection{Noise-predictor architecture}

Our noise-predicting network $\denoiser$ uses an architecture similar to CSDI~\cite{tashiro_csdi_2021}, which has shown promising results for time-series imputation~\cite{tashiro_csdi_2021}, including human pose forecasting~\cite{saadatnejad_generic_2022}. CSDI inherits some design choices from DiffWave~\cite{kong2020diffwave}, an architecture used for audio synthesis.

Our design has 16 residual blocks, each containing two GCN layers with batch normalization, ReLU activation, and dropout, followed by a graph non-local layer~\cite{zhao_semantic_2019}. We used pre-aggregated graph convolutions with decoupled self-connections~\cite{liu_comprehensive_2020} and 64-dimensional embeddings. The two graph-convolutional layers in each block replace the two transformer layers from CSDI~\cite{tashiro_csdi_2021, saadatnejad_generic_2022}, allowing for much faster training and inference. Like in CSDI transformer layers, those two graph layers alternate between time-wise (independent convolutions for each feature, carried across time) and feature-wise (independent for each time step, carried across features). In our design, the skeleton connectivity of the human pose is always intrinsically exploited through the graph convolution connectivity. The blocks are connected through gated activation units to produce residuals, and we use the same inter-block and block-output connectivity as CSDI.

\subsection{Training DiffHPE} \label{sec:training}

We train DiffHPE as a standard DDPM~\cite{ho_denoising_2020}. For a given example pair ${(\xdd, \xtd)}$:
\begin{enumerate}[leftmargin=*,itemsep=0pt]
    \item We sample a time-step $t \sim \mathcal{U}(\{1, \dots, T\})$;
    \item We sample  ${\varepsilon \in \R^{L \times J \times 3}} \sim \mathcal{N}(0, \mathbf{I})$ and use it to derive the noisy 3D data $\xtd_t$;
    \item We compute $\edd=\f{h}_*(\xdd)$ and $\etd_t=\enc(\xtd_t)$, and use them to predict the noise $\hat{\varepsilon}=\denoiser(\etd_t, \edd, t)$;
    \item We compare $\varepsilon$ and $\hat{\varepsilon}$ with a $L_2$-loss, back-propagated to $\denoiser$ and $\enc$ to learn the model parameters $\theta$.
\end{enumerate}
The whole scheme is illustrated  in \autoref{fig:training}. Steps 1--2 correspond to eqs.~(\ref{eq:fwd-alpha}--\ref{eq:fwd-proc2}), which allow for efficient sampling of training steps. We apply the steps above for all examples in the training dataset, repeating the epochs until convergence.

\subsection{DiffHPE during inference}

Inference follows a standard DDPM procedure (Fig.\,\ref{fig:inference}):
\begin{enumerate}[leftmargin=*,itemsep=0pt]
    \item We sample  $\xtd_T \in \R^{L \times J \times 3} \sim \mathcal{N}(0, \mathbf{I})$;
    \item We compute $\edd=\f{h}_*(\xdd)$;
    \item For all time steps $t$, in reverse sequence from $T$ to $1$, we:
    \begin{enumerate}
        \item Compute $\etd_t=\enc(\xtd_t)$, and use it to predict the noise $\hat{\varepsilon}_t=\denoiser(\etd_t, \edd, t)$;
        \item Compute $\mu_t$ from $\xtd_t$ and $\hat{\varepsilon}_t$, following eq.~\eqref{eq:ddpm-cond-mixste};
        \item Sample $\xtd_{t-1} \sim \mathcal{N}(\mu_t, \Sigma_t)$\,,
    \end{enumerate}
\end{enumerate}
where steps 3.b and 3.c depend on non-learned constants computed from the variance schedule.

The process is obviously non-deterministic, allowing, from a single $\xdd$ input, to sample many 3D poses $\xtd$. We may see those samples as a parameterizable number $H$ of hypotheses for the estimated pose, which are aggregated into a final prediction.

In the literature on generative techniques, the most common ``aggregation'' technique is simply picking the best pose (the one closest to the ground truth), thus assessing an upper-bound performance. That is obviously unrealistic, and unfair if the comparison includes deterministic techniques. The simplest solution, which we adopt here, is to average the samples. More sophisticated strategies, such as joint-wise reprojection-based aggregation~\cite{shan2023diffusion} could be used in situations where the intrinsic and extrinsic camera parameters are available to reproject the sampled 3D poses onto the 2D image space.

\renewcommand{\arraystretch}{1.1}
\begin{table}[t]
\resizebox{\columnwidth}{!}{%
\begin{tabular}{rcccc}
\toprule
                   & Batch size & Learning rate & Dropout & Epochs \\ \midrule
MixSTE\cite{zhang_mixste_2022}             & $1024$              & $4.0 \times 10^{-5}$              & $0.10$            & $~\,500$           \\
DiffHPE-2D & $~\,200$               & $8.0 \times 10^{-4}$              & $0.03$           & $1000$          \\
DiffHPE-Wrapper        & $~\,200$               & $2.7 \times 10^{-4}$            & $0.27$           & $1000$          \\ \bottomrule
\end{tabular}%
}
\smallskip\caption{
    \textbf{Training hyperparameters used in experiments}. When possible, we used hyperparameters reported in \cite{zhang_mixste_2022} for MixSTE, while the diffusion models' were tuned with random search.
}
\label{tab:hyperparams}
\end{table}

\renewcommand{\arraystretch}{1.}
\begin{table*}[ht]
\centering
\begin{tabular}{rccccccc}
\toprule
 &
  Diffusion &
  Condition &
  $H$ &
  \begin{tabular}[c]{@{}c@{}}Sequence\\ length\end{tabular} &
  \begin{tabular}[c]{@{}c@{}}MPJPE\\~ [mm]$\downarrow$\end{tabular} &
  \begin{tabular}[c]{@{}c@{}}Symmetry\\ gap [mm]$\downarrow$\end{tabular} &
  \begin{tabular}[c]{@{}c@{}}Segments length\\ std [mm]$\downarrow$\end{tabular}\\ \midrule
MixSTE \cite{zhang_mixste_2022}$^\dagger$            & \xmark & N/A & 1 & 27 & {\ul 51.8} & -   & -   \\
MixSTE$^\ddagger$             & \xmark & N/A & 1 & 27 & 54.8 & 18.2 & 5.2 \\
DiffHPE-2D (ours) & \cmark & $\mathbf{x}^{2D}$ & 5 & 27 & {\ul 51.8} &    ~\,\textbf{7.0} &   \textbf{2.1}  \\
DiffHPE-Wrapper (ours)        & \cmark & $\edd$ & 5          & 27 & \textbf{51.2} & {\ul 12.4} & {\ul 3.2} \\ \bottomrule
\end{tabular}%
\smallskip\caption{
    \textbf{Comparison between MixSTE and DiffHPE}. DiffHPE-2D and MixSTE$\ddagger$ act as an ablation (respectively of the conditioning on pre-trained MixSTE, and of the diffusion refinement) for DiffHPE-Wrapper. $H$ is the number of samples used to average the predicted pose. MixSTE$^\dagger$ is verbatim from \cite{zhang_mixste_2022}, while MixSTE$^\ddagger$ is reproduced from the same official code, but without the hand-tuned weighted loss and test-time augmentation, for better comparison. (\textbf{Best} and \uline{second best} results as indicated.) 
    }
\label{tab:main-tab}
\end{table*}

\section{Experimental setting}
\subsection{Dataset and metrics}

We use the Human\,3.6M dataset~\cite{ionescu_human36m_2014}, the most widely used dataset for 3D human pose estimation. It contains 3.6 million images of 7 actors performing 15 different actions.
Both 2D and 3D ground-truth keypoints positions are available for 4 static camera viewpoints.

Following previous works~\cite{zhang_mixste_2022, gong_diffpose_2022, zhao_semantic_2019}, we train our models on subjects S1, S5, S6, S7, S8, and test on subjects S9 and S11.

We use the mean-per-joint-position error (MPJPE), with a 17-joint skeleton, as the main metric, following most previous works \cite{gong_diffpose_2022, zhang_mixste_2022, zhao_semantic_2019, martinez_simple_2017}. We compute the metric under protocol \#1, \ie, after translating the predicted root joint to its correct position.

\subsection{Implementation details}

\myparagraph{Models.} We used MixSTE~\cite{zhang_mixste_2022}, pre-trained on the same Human\,3.6M data~\cite{ionescu_human36m_2014}, as the frozen lifting feature extractor $\f{h}_*$. Remark that on all experiments, the 2D input to the models, during training and test, are 2D keypoints predicted from a CPN~\cite{chen2018cascaded} model, trained on the same data. That is important because sometimes results are reported on ground-truth 2D annotations, which leads to more optimistic results. All models were implemented in \textsc{PyTorch}~\cite{paszke2019pytorch}.

\myparagraph{Training.} We trained the diffusion models for up to $1000$ epochs using the Adam optimizer~\cite{kingma2014adam} with default parameters $\beta_1=0.9$, $\beta_2=0.999$, and weight decay $=10^{-6}$. We tuned the learning rate and the dropout for all diffusion models with a random search~\cite{bergstra_random_2012} implemented in \textsc{Hydra}~\cite{Yadan2019Hydra} and \textsc{Optuna}~\cite{optuna_2019} (\autoref{tab:hyperparams}). We used a plateau learning-rate scheduler with a factor of 2 and patience of 50 epochs. We set the number of diffusion steps $T=50$ for all experiments.

Following current practice in literature and associated official code repositories that benchmark on Human\,3.6M, we used the test performance both to select the hyperparameters and to pick the best checkpoint during training.

We retrained the supervised MixSTE baseline with its original parameters~\cite{zhang_mixste_2022} and following the official code repository.\footnote{\url{https://github.com/JinluZhang1126/MixSTE}} We trained MixSTE for up to 500 epochs, significantly more than the 120 epochs reported in the paper, to ensure it had fully converged.

\section{Results}

\subsection{Diffusion improves pre-trained lifting models}
\label{sec:results-improves-lifting}

DiffHPE-Wrapper improves upon MixSTE, demonstrating its ability to refine the pre-trained model's predictions (\autoref{tab:main-tab}). In the same table, an ablation comparing both DiffHPE models showcases the compromise of using a strong baseline such as MixSTE as conditioning. On the one hand, the combined model DiffHPE-Wrapper has the best accuracy performance; on the other hand, the ``purer'' DiffHPE-2D model has the best indicators for pose symmetry and temporal coherence. While DiffHPE-Wrapper considerably mitigates those indicators for MixSTE, DiffHPE-2D, working \textit{ab initio}, does not inherit MixSTE biases. The coherence results are analyzed in-depth in \autoref{sec:results-improves-coherence}, together with a detailed explanation for the metrics employed.

We report two performance values for MixSTE: one copied \emph{verbatim} from the paper and another obtained by reproducing the technique in training and testing conditions identical to DiffHPE, allowing for a fairer comparison. All experiments in this table used a video length of 27 frames.

\autoref{tab:meta-analysis}, a compilation of recent experiments from literature on longer sequences of 243 video frames, reveals the same trends for accuracy as \autoref{tab:main-tab}. Unfortunately, bilateral and temporal coherence has not received much attention, preventing to evaluate trends for those metrics in the same compilation.

\begin{table}[ht]
\resizebox{\columnwidth}{!}{%
\begin{tabular}{rccccc}
\toprule
 &
  Diffusion &
  Condition &
  $H$ &
  \begin{tabular}[c]{@{}c@{}}Seq.\\ length\end{tabular} &
  \begin{tabular}[c]{@{}c@{}}MPJPE\\ ~[mm]$\downarrow$\end{tabular} \\ \midrule
MixSTE \cite{zhang_mixste_2022} & \xmark & N/A & 1  & 243 & 40.9          \\
D3DP \cite{shan2023diffusion} & \cmark & $\mathbf{x}^{2D}$ & 20  & 243 & 39.5          \\
DiffPose \cite{gong_diffpose_2022} & \cmark & $\edd$ & 5 & 243 & \textbf{36.9} \\ \bottomrule
\end{tabular}%
}
\smallskip\caption{
    \textbf{Compilation of state-of-art human-pose lifting models}. We observe the same trends as in \autoref{tab:main-tab}, \ie, the diffusion models (D3DP and DiffPose) outperform the deterministic model (MixSTE), and DiffPose (which uses diffusion~+~MixSTE) outperforms D3DP, even with the latter using the camera parameters to aggregate 3D poses in a more sophisticated way. (Best in \textbf{bold}.)
}
\label{tab:meta-analysis}
\end{table}

\subsection{Diffusion improves lifting under occlusions} \label{sec:occ-res}

Given the success of diffusion models in solving tasks such as inpainting~\cite{saharia2022palette, lugmayr2022repaint} and time-series imputation~\cite{tashiro_csdi_2021}, they appear as natural candidates to address occlusions, a major challenge in human pose analysis.
Indeed, they demonstrate that ability on human pose forecasting under occlusions~\cite{saadatnejad_generic_2022}. Here, we evaluate 3D human pose lifting under the same challenging occlusion patterns proposed in~\cite{saadatnejad_generic_2022}, namely:
\begin{enumerate}[itemsep=1mm]
    \item \textbf{Random}: any 2D keypoint in any frame may be omitted with an equal probability $p=0.2$ ;
    \item \textbf{Random leg and arm}: any frame has a probability $p=0.4$ that all left arm and right leg keypoints are omitted;
    \item \textbf{Consecutive leg}: a sequence of 10 consecutive frames ($\sim 40\%$ of total sequence length, picked uniformly at random) has all right leg keypoints omitted;
    \item \textbf{Consecutive frames}: a sequence of 5 consecutive frames ($\sim 20\%$ of total sequence length, picked uniformly at random) has all keypoints completely omitted.
\end{enumerate}
Omitting a keypoint corresponds to setting its value to 0.

We trained and evaluated three models (one supervised baseline MixSTE and two versions of DiffHPE-Wrapper) on all occlusion patterns, including no occlusions. Of the two DiffHPE-Wrapper models, one was conditioned on a vanilla MixSTE (trained without occlusions), and the other was conditioned on a MixSTE trained with the occlusions. The results (\autoref{tab:occ}) show that the occlusion-trained DiffHPE beats the occlusion-trained MixSTE on all patterns. Surprisingly, the DiffHPE-Wrapper with vanilla MixSTE is competitive with the occlusion-trained MixSTE for all but the Random pattern.

\renewcommand{\arraystretch}{1.}
\begin{table}[ht]
\resizebox{\columnwidth}{!}{%
\begin{tabular}{lccc}
\toprule
                                      & \multicolumn{1}{l}{MixSTE\cite{zhang_mixste_2022}} & \multicolumn{2}{l}{DiffHPE-Wr.}       \\
With diffusion               & \xmark                          & \cmark             & \cmark \\
Conditioning trained w/ occ. & N/A                        & \xmark             & \cmark                     \\ \midrule
No occlusion                          & {\ul 54.8}                 & \textbf{51.2} & \textbf{51.2}         \\
Random                                & {\ul 54.5}                 & 60.3          & \textbf{53.2}         \\
Random leg and arm                    & {\ul 54.4}                 & 55.2          & \textbf{53.0}         \\
Consecutive leg                       & 55.1                       & {\ul 54.1}    & \textbf{52.6}         \\
Consecutive frames                    & 55.8                       & \textbf{52.2} & {\ul 53.5}            \\ \bottomrule
\end{tabular}%
}
\smallskip\caption{\textbf{Impact of occlusions}.
    Performance (MPJPE in mm) with and without diffusion under different \emph{known} occlusion patterns. (\textbf{Best} and \uline{second best} results as indicated.)
}
\label{tab:occ}
\end{table}

\subsection{Diffusion improves robustness to occlusion-pattern misspecification}

Continuing from \autoref{sec:occ-res}, note that training with simulated occlusions may be interpreted as a form of data augmentation. Yet, occlusions found during test time might differ from those used for training, raising the question of robustness to such domain gaps.

\begin{figure}[ht]
    \centering
    \includegraphics[width=0.93\columnwidth]{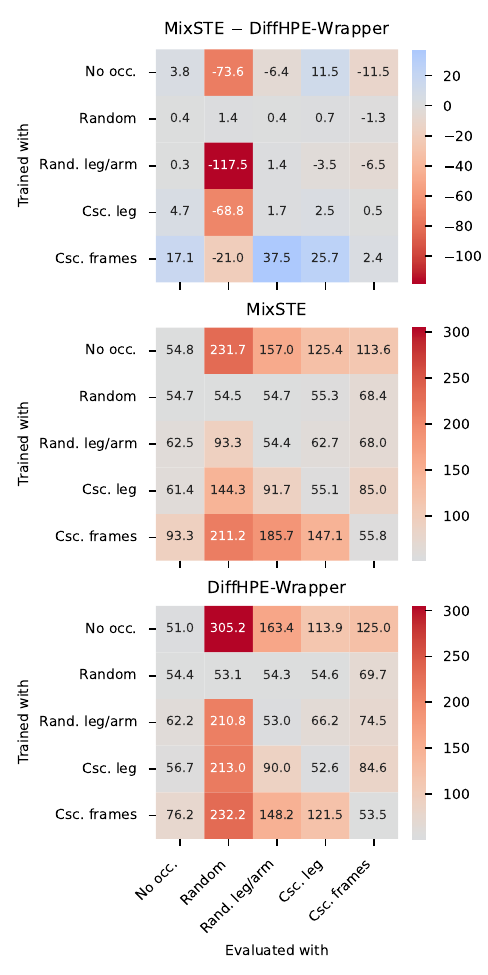}
    \caption{\textbf{Impact of train/test discrepancy between occlusion patterns}.  
    Cross-domain accuracy (MPJPE in mm) with and without diffusion under different occlusion patterns.
    Rows correspond to occlusion patterns at training, and columns indicate occlusions at test time. The bottom two matrices show the MJPE of each method (lower is better), and the top matrix, their difference, with positive values (blue) favoring DiffHPE-Wrapper.
    }
    \label{fig:occ-matrix-mpjpe}
\end{figure}

To investigate this question, we evaluate both occlusion-trained models (MixSTE and DiffHPE-Wrapper conditioned on it) in a cross-domain setting where training and test occlusion patterns do not necessarily match.
Those results appear in \autoref{fig:occ-matrix-mpjpe}. We find that, in general, an occlusion pattern misspecification hurts the performance of both models. The performance contrast showcases interesting differences between the two models, with the diffusion model being more robust to most cases, but being particularly vulnerable to misspecification when testing on the fully Random case, suggesting the importance of structured patterns for generative modeling. On the other hand, diffusion is particularly robust when training on missing consecutive frames.

\subsection{Diffusion improves pose coherence}
\label{sec:results-improves-coherence}

The results in \autoref{tab:main-tab} already suggest that diffusion improves two aspects of the estimated poses coherence:

\myparagraph{Symmetry.} We evaluate the sagittal symmetry of predicted skeletons, \ie, segments having the same length on both sides of the body, with the average absolute difference between the length of left- and right-side segments predicted:
\begin{equation}
    \frac{1}{K L \,S_{\text{left}}} \sum_{i=1}^K \sum_{l=1}^L \sum_{s=1}^{S_{\text{left}}} |y_{i,l,s}- y_{i, l, \tau(s)}| \,,
\end{equation}
where $K$ is the number of test sequences, $L$ is the sequence length, $S_{\text{left}}$ is the number of segments on the left-side, $\tau$ maps left-side segments indices to their right-side counterpart, and $y_{i,l,s}$ is the segment's length, given by the Euclidean distance between its two joints.

\myparagraph{Temporal coherence.} We evaluate time coherence, \ie, segment lengths not varying within a sequence, using the average of the time-wise standard deviations for each predicted segment series:
\begin{gather}
    \frac{1}{K S} \sum_{i=1}^K \sum_{s=1}^{S} \sqrt{ \frac{1}{L} 
            \sum_{l=1}^L (y_{i,l,s} - \bar{y}_{i, s})^2 } \,, 
\end{gather}
where ${\bar{y}_{i, s} = \frac{1}{L} \sum_{l=1}^L y_{i,l,s}}$ and $S$ is the total number of segments.

\begin{figure}[ht]
    \centering
    \includegraphics[width=0.94\columnwidth]{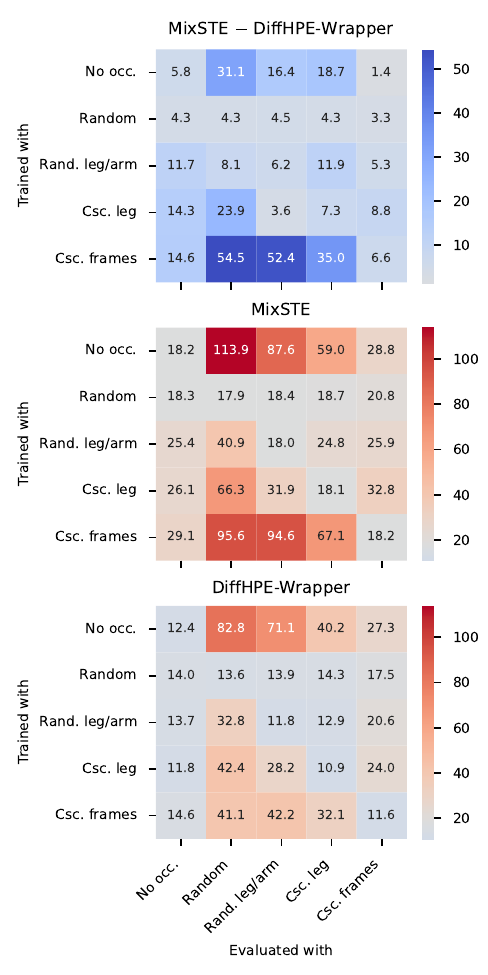}
    \caption{
        \textbf{Impact of occlusions on sagittal symmetry of predicted poses}. Average symmetry gap (in mm) with and without diffusion across different training and test occlusion patterns. The two bottom matrices show gap for the two methods (lower is better), and the top matrix, their difference, with positive values (blue) favoring DiffHPE-Wrapper.
    }
    \label{fig:occ-matrix-sym}
\end{figure}

\begin{figure}[ht]
    \centering
    \includegraphics[width=0.94\columnwidth]{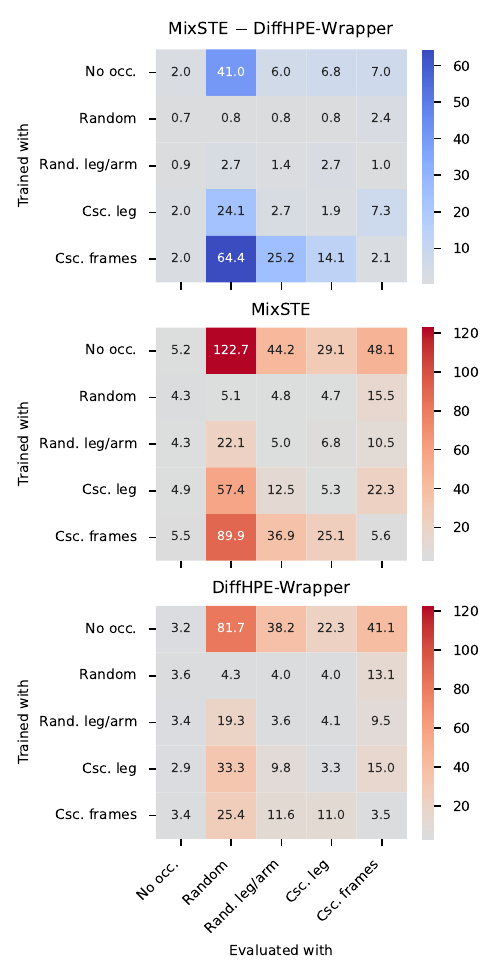}
    \caption{
        \textbf{Impact of occlusions on temporal consistency of segments' length}. Average time-wise standard deviation of predicted segments lengths (in mm) across different training and test occlusion patterns. The two bottom matrices show the deviation for the two methods (lower is better), and the top matrix, their difference, with positive values (blue) favoring DiffHPE-Wrapper.
    }
    \label{fig:occ-matrix-std}
\end{figure}

Improvements similar to \autoref{tab:main-tab} appear
in \autoref{fig:occ-matrix-sym} and \autoref{fig:occ-matrix-std}, where DiffPose-Wrapper improves the coherence across all combinations of training and test occlusion-patterns (following the protocol of the previous section), including those where the accuracy results are not
improved (\textit{cf}. \autoref{fig:occ-matrix-mpjpe}). That latter remark is interesting, as it suggests diffusion's ability to improve the coherence of estimations
independently of ``raw'' accuracy gains, maybe through learning subtler hints about the data distribution.

\section{Conclusion}

We have investigated how to use diffusion models for 3D human pose lifting effectively. We show that, although diffusion models work well when used directly (DiffHPE-2D), associating them with state-of-the-art supervised models leads to even better results (DiffHPE-Wrapper). Our results demonstrate that diffusion models not only lead to more accurate predictions, but also
to more time-coherent poses, which are also more compatible with the symmetry of the human body.

In future work, we plan to extend our analyses to longer sequences and find ways to mitigate the greater computational burden of diffusion, which remains its main drawback compared to classical approaches, especially during inference.

We hope this work will serve as a catalyst, showcasing the potential of diffusion models in predictive tasks and raising questions that may inspire and foster new research in this exciting area.

\subsection*{Acknowledgments}

We are grateful to Saeed Saadatnejad for the source code of TCD~\cite{saadatnejad_generic_2022}. This work was granted access to the HPC resources of IDRIS under the allocation 2023-AD011014073 made by GENCI.

\clearpage
{\small
\bibliographystyle{ieee_fullname}
\bibliography{pose-estimation}
}

\end{document}